\documentclass[11pt]{article}
\usepackage[utf8]{inputenc}
\usepackage[T1]{fontenc}
\usepackage[margin=1in]{geometry}
\usepackage{amsmath,amssymb,amsfonts}
\usepackage{booktabs,tabularx,array,longtable,multirow}
\usepackage{graphicx}
\usepackage{microtype}
\usepackage{enumitem}
\usepackage{xcolor}
\usepackage{hyperref}
\usepackage{tikz}
\usetikzlibrary{arrows.meta,positioning,calc}
\hypersetup{colorlinks=true,linkcolor=blue,citecolor=blue,urlcolor=blue}
\newcolumntype{Y}{>{\raggedright\arraybackslash}X}
\title{A Proactive EMR Assistant for Doctor-Patient Dialogue:\\Streaming ASR, Belief Stabilization, and Preliminary Controlled Evaluation}
\author{
Zhenhai Pan$^{1}$ \quad Yan Liu$^{1}$ \quad Jia You$^{1}$\\[0.6em]
$^{1}$The Hong Kong Polytechnic University, Hong Kong
}
\date{}
\begin{document}
\maketitle

\begin{abstract}
Most dialogue-based electronic medical record (EMR) systems still behave as passive pipelines: transcribe speech, extract information, and generate the final note after the consultation. That design improves documentation efficiency, but it is insufficient for proactive consultation support because it does not explicitly address streaming speech noise, missing punctuation, unstable diagnostic belief, objectification quality, or measurable next-action gains. We present an end-to-end proactive EMR assistant built around streaming speech recognition, punctuation restoration, stateful extraction, belief stabilization, objectified retrieval, action planning, and replayable report generation. The system is evaluated in a preliminary controlled setting using ten streamed doctor-patient dialogues and a 300-query retrieval benchmark aggregated across dialogues. The full system reaches state-event F1 of 0.84, retrieval Recall@5 of 0.87, and end-to-end pilot scores of 83.3\% coverage, 81.4\% structural completeness, and 80.0\% risk recall. Ablations further suggest that punctuation restoration and belief stabilization may improve downstream extraction, retrieval, and action selection within this pilot. These results were obtained under a controlled simulated pilot setting rather than broad deployment claims, and they should not be read as evidence of clinical deployment readiness, clinical safety, or real-world clinical utility. Instead, they suggest that the proposed online architecture may be technically coherent and directionally supportive under tightly controlled pilot conditions. The present study should be read as a pilot concept demonstration under tightly controlled pilot conditions rather than as evidence of clinical deployment readiness or clinical generalizability.
\end{abstract}

\section{Introduction}
Clinical note generation from doctor-patient dialogue has advanced rapidly in recent years \cite{molenaar2020medical,abacha2023empirical,abacha2023mediqa}. Yet most systems remain fundamentally passive: they summarize or structure the encounter after the dialogue has already happened. A proactive assistant must operate differently. It must process streaming input, maintain an explicit state of what is currently known, identify what remains missing, retrieve clinically relevant support, and suggest the next best action before the interaction has ended.

This requirement introduces several engineering bottlenecks that are often under-discussed in method-centric papers. First, speech input is streaming and noisy. Missing punctuation is not just a cosmetic issue; it degrades boundary recovery, evidence grouping, negation scope, and action triggering. Second, belief estimates derived from language-model outputs are often unstable; naive pseudo-probabilities can oscillate sharply and lead to erratic next-action recommendations. Third, retrieval quality depends not only on embeddings, but also on how documents are parsed, objectified, and anchored. Finally, evaluation should not stop at note quality: a proactive system should also be judged by how efficiently and safely it closes information gaps and reaches a target state.

The core methodological formulation --- stateful extraction, sequential belief updating, hybrid retrieval, and POMDP-lite action planning --- is described in the companion methods paper \cite{pan2026methods}. Here we focus on system realization. We show how to convert that formulation into an auditable online pipeline and evaluate it in a preliminary controlled setting.

Our contributions are fourfold. First, we present an end-to-end system prototype for proactive dialogue-based EMR assistance, spanning audio input, punctuation restoration, stateful extraction, belief stabilization, retrieval, action selection, report generation, and replay. Second, we define the engineering interfaces that make the online stack auditable under streaming input. Third, we report targeted ablations for punctuation restoration and belief stabilization. Fourth, we provide a preliminary controlled evaluation with explicit raw-count denominators so that later large-scale studies can be compared against a stable systems baseline.

\section{System Overview}
\subsection{Pipeline structure}
The system is designed as a layered consultation-support pipeline with eight online stages: audio acquisition, ASR, punctuation restoration, stateful extraction, belief update, hybrid retrieval, action planning, and report generation with replay support. Unlike passive note generators, this architecture is explicitly turn-aware. At each step, the system maintains a structured current state, compares it against a dynamic goal state, derives typed gap signals, and chooses the next action accordingly.

\begin{figure}[t]
\centering
\begin{tikzpicture}[node distance=0.55cm and 0.45cm, every node/.style={font=\small}]
\tikzstyle{box}=[draw, rounded corners, minimum height=0.75cm, minimum width=2.0cm, align=center]
\node[box] (a) {Audio};
\node[box, right=of a] (b) {ASR};
\node[box, right=of b] (c) {Punctuation\\restoration};
\node[box, right=of c] (d) {Stateful\\extraction};
\node[box, right=of d] (e) {Belief\\update};
\node[box, below=0.9cm of d] (f) {Hybrid\\retrieval};
\node[box, right=of f] (g) {Action\\planning};
\node[box, right=of g] (h) {EMR / replay};
\draw[-{Latex[length=2mm]}] (a) -- (b);
\draw[-{Latex[length=2mm]}] (b) -- (c);
\draw[-{Latex[length=2mm]}] (c) -- (d);
\draw[-{Latex[length=2mm]}] (d) -- (e);
\draw[-{Latex[length=2mm]}] (d) -- (f);
\draw[-{Latex[length=2mm]}] (e) -- (g);
\draw[-{Latex[length=2mm]}] (f) -- (g);
\draw[-{Latex[length=2mm]}] (g) -- (h);
\end{tikzpicture}
\caption{Online execution pipeline of the proactive EMR assistant.}
\label{fig:pipeline}
\end{figure}
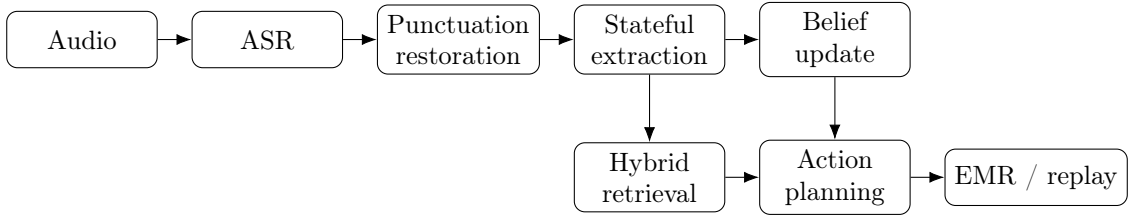

\subsection{Offline time and online time}
A useful implementation distinction is between offline time and online time. Offline time includes terminology initialization, document parsing, block reconstruction, objectification, graph construction, and index building. These steps prepare the knowledge side of the system. Online time begins once a consultation starts. Streaming audio is transcribed, punctuation is restored, utterances are converted into stateful events, belief is updated, gaps are identified, and actions are selected.

\subsection{Medical instantiation}
Although the platform can be generalized to education, law, and finance, the implementation in this paper is instantiated as a medical consultation package. This package defines domain-specific schema, field meta-models, terminology normalization rules, role mapping, retrieval profiles, graph relations, action policies, report templates, and risk priorities.

\section{Key Engineering Components}
\subsection{Streaming ASR and punctuation restoration}
In the current project baseline, ASR output can arrive as unpunctuated spans such as: ``started yesterday chest feels tight worse when climbing stairs gets a little better after sitting left shoulder sometimes sore.'' For downstream structured processing, the lack of punctuation is not cosmetic: it directly affects sentence boundaries, local evidence grouping, temporal interpretation, negation scope, and action triggers. We therefore insert a punctuation restoration layer immediately after ASR and before stateful extraction, following the general intuition of punctuation-recovery work while adapting it to streamed clinical dialogue \cite{tilk2016punctuation,sunkara2020medicalasr}. The design goal is not perfect written-language punctuation, but clinically useful boundary restoration.

Let a streaming token sequence be $x_1,x_2,\ldots,x_n$. For a candidate boundary after token $x_i$, we compute a boundary score
\begin{equation}
b_i=\alpha_1 p_i + \alpha_2 \ell_i + \alpha_3 r_i + \alpha_4 q_i,
\end{equation}
where $p_i$ is a pause-derived score, $\ell_i$ a lexical cue score, $r_i$ a role-transition cue score, and $q_i$ a prosodic or confidence-derived score when available. The corresponding boundary probability is
\begin{equation}
\hat{y}_i = \sigma(b_i)=\frac{1}{1+\exp(-b_i)}.
\end{equation}

\subsection{Belief stabilization}
Raw language-model outputs are useful semantic cues, but not reliable calibrated clinical posteriors. We therefore add an explicit stabilization layer rather than using raw pseudo-probabilities directly \cite{guo2017calibration,angelopoulos2023conformal}. Given raw logits $z_i$ over candidate hypotheses, we first apply temperature scaling
\begin{equation}
p_i^{(T)} = \frac{\exp(z_i/T_t)}{\sum_j \exp(z_j/T_t)},
\end{equation}
where $T_t$ is dynamically adjusted according to speech quality, rule confidence, and recent volatility. We then fuse prior stabilized belief, rule evidence, retrieval evidence, and model outputs:
\begin{equation}
\hat{b}_t(h)=\alpha\tilde{b}_{t-1}(h)+\beta s_{\mathrm{rule}}(h)+\gamma s_{\mathrm{retrieval}}(h)+\delta s_{\mathrm{llm}}(h).
\end{equation}
To suppress sharp oscillations, we apply exponential smoothing
\begin{equation}
\tilde{b}_t(h)=\lambda \tilde{b}_{t-1}(h)+(1-\lambda)\hat{b}_t(h), \qquad \lambda\in[0.7,0.9].
\end{equation}
Candidate actions are ranked using a stabilized form of expected information gain,
\begin{equation}
\mathrm{EIG}(a)=H_0-\bar{H}_a-\eta V_a,
\end{equation}
where $H_0$ is the current entropy, $\bar{H}_a$ the Monte Carlo estimate of post-action average entropy, and $V_a$ the variance across simulated future observations.

\subsection{Objectification and hybrid retrieval}
The knowledge subsystem operates on parsed and objectified content rather than plain text chunks alone, extending retrieval-augmented processing beyond chunk-only matching \cite{lewis2020rag}. Documents are routed per page to native-text extraction, OCR, layout-aware OCR, or fallback OCR, drawing on document-understanding components such as layout-aware pretraining, OCR-free parsing, table extraction, and lightweight OCR pipelines \cite{huang2022layoutlmv3,kim2021donut,smock2022pubtables,li2022ppocrv3}. Parsed content is reconstructed into semantically coherent blocks and then mapped into medical objects such as symptom units, exam units, diagnosis units, risk-rule units, and case-summary objects. Each object is assigned an anchor
\begin{equation}
a(o_i)=(\mathrm{document\_id},\mathrm{page\_no},\mathrm{block\_id},\mathrm{span}),
\end{equation}
which supports traceability and replay.

\subsection{Report generation and replay}
The system produces two synchronized outputs: a structured EMR-oriented output for current use and a replayable trace for audit, debugging, and research analysis. The trace stores extracted events, state transitions, belief evolution, retrieval outputs, candidate actions, selected actions, and references to supporting anchors.

\section{Experimental Setup}
\subsection{Preliminary controlled setting}
We evaluate the system on ten protocol-driven doctor-patient dialogues designed to cover representative consultation scenarios with meaningful uncertainty and action choices. These include chest discomfort, upper abdominal pain, and other risk-sensitive cases where the system must distinguish routine information gathering from risk-closing inquiry. Each case contains a gold case card, gold stateful events, gold target-state requirements, expected risk signals, and preferred next-action patterns.

To exercise the online stack under streamed input, each scripted dialogue was read aloud by two speakers and replayed through the system as a real audio stream with preserved turn order, pauses, and disfluency-like hesitations. ASR, boundary recovery, and punctuation restoration are therefore evaluated on actual ASR outputs from recorded audio streams rather than on manually de-punctuated text. Across the ten dialogues, the end-to-end audit includes 180 gold information items, 140 structure-critical record slots, and 60 gold risk items. Retrieval is evaluated on a 300-query benchmark aggregated across dialogues. The 300 retrieval queries are protocol-defined unresolved-state prompts derived from the gold schema of the ten pilot cases, rather than naturally occurring user queries from independent encounters.

\subsection{Baselines}
We compare four configurations: (A) direct generation from dialogue without proactive control, (B) a chunk-only retrieval-augmented pipeline, (C) a rule-template interactive pipeline, and (D) the full system with streaming ASR, punctuation restoration, stateful extraction, stabilized belief, hybrid retrieval, and action planning. All interactive configurations use the same evaluation cases, target-state schema, and underlying knowledge corpus; differences arise from state tracking, retrieval granularity, belief handling, and action policy rather than from access to different evidence sources.

\subsection{Metrics}
We evaluate four layers of performance.

\textbf{Extraction layer.} Let $TP$, $FP$, and $FN$ denote true-positive, false-positive, and false-negative stateful events relative to gold state-event annotations. Precision, recall, and F1 are computed in the standard way.

\textbf{Retrieval layer.} We report Recall@5, MRR@5, and nDCG@5 on the 300-query benchmark. Raw hit counts are reported alongside percentages wherever possible.

\textbf{Action layer.} For the stabilization ablation, \emph{volatility} is defined as the mean adjacent-turn $L_1$ belief change,
\begin{equation}
\mathrm{Volatility}=\frac{1}{T-1}\sum_{t=2}^{T}\|\tilde{b}_t-\tilde{b}_{t-1}\|_1.
\end{equation}
We also report EIG variance and the number of wrong actions, defined as turns on which the selected action disagrees with the gold preferred next-action type or fails to choose a required risk-closing action. Action Utility is reported as a supportive internal control metric derived from the utility formulation in the companion methods paper, and it should be interpreted together with Risk Recall, Redundancy, and $T_{\mathrm{goal}}$ rather than as a standalone benchmark metric.

\textbf{End-to-end layer.} Let $Y$ denote the set of gold target information items, $\hat{Y}$ the subset correctly covered by the system, $S$ the set of required structural slots, $S^+$ the subset filled with semantically correct values, and $R$ the set of gold critical-risk items. We report Coverage $=|\hat{Y}|/|Y|$, Structural Completeness $=|S^+|/|S|$, Risk Recall $=|\hat{R}|/|R|$, and $T_{\mathrm{goal}}$, the first turn at which mandatory goal-state slots and predefined risk checks have been satisfied.

\section{Results}
\subsection{End-to-end pilot summary}
Table~\ref{tab:overallsys} summarizes the overall pilot results. The full system covers 150/180 gold information items, fills 114/140 structural slots correctly, surfaces 48/60 risk items, and produces 15 redundant prompts out of 95 total prompts. Table~\ref{tab:overallsys} is repeated only for reader orientation and to anchor the systems discussion; the system-specific evidence of this paper begins with Tables~\ref{tab:extract}--\ref{tab:latency}. The detailed methodological formulation is presented in the companion paper \cite{pan2026methods}.

\begin{table}[t]
\centering
\caption{End-to-end pilot results in the controlled setting. Higher is better for Coverage, Structural Completeness, Action Utility, and Risk Recall. Lower is better for Redundancy and $T_{\mathrm{goal}}$. Action Utility is a supportive internal control metric and should be interpreted together with Risk Recall, Redundancy, and $T_{\mathrm{goal}}$ rather than as a standalone benchmark metric. N/A indicates metrics not applicable to the non-interactive baseline.}
\label{tab:overallsys}
\small
\begin{tabularx}{\textwidth}{Y>{\centering\arraybackslash}p{0.11\textwidth}>{\centering\arraybackslash}p{0.12\textwidth}>{\centering\arraybackslash}p{0.10\textwidth}>{\centering\arraybackslash}p{0.11\textwidth}>{\centering\arraybackslash}p{0.10\textwidth}>{\centering\arraybackslash}p{0.09\textwidth}}
\toprule
Method & Coverage (\%) & Structural Completeness (\%) & Action Utility & Redundancy (\%) & Risk Recall (\%) & $T_{\mathrm{goal}}$ \\
\midrule
Baseline A: Direct generation & 55.6 & 59.3 & 0.41 & N/A & 30.0 & N/A \\
Baseline B: Chunk-only RAG & 69.4 & 71.4 & 0.54 & 24.8 & 60.0 & 6.8 \\
Baseline C: Rule-template interaction & 77.8 & 78.6 & 0.61 & 28.4 & 70.0 & 7.4 \\
Full system & 83.3 & 81.4 & 0.69 & 15.8 & 80.0 & 5.8 \\
\bottomrule
\end{tabularx}
\end{table}

\subsection{Extraction-layer results}
Table~\ref{tab:extract} reports micro-averaged precision, recall, and F1 over normalized stateful-event annotations. Baseline A does not maintain explicit state events and is therefore not directly comparable on this layer. For the full system, a consistent raw-count audit corresponds to $TP=91$, $FP=16$, and $FN=19$, which is consistent with micro-level precision of 0.85, recall of 0.83, and F1 of 0.84.

\begin{table}[t]
\centering
\caption{Extraction results on gold stateful-event annotations.}
\label{tab:extract}
\begin{tabular}{lccc}
\toprule
Method & Precision & Recall & F1 \\
\midrule
Baseline B: Chunk-only RAG & 0.76 & 0.72 & 0.74 \\
Baseline C: Rule-template interaction & 0.81 & 0.79 & 0.80 \\
Full system & 0.85 & 0.83 & 0.84 \\
\bottomrule
\end{tabular}
\end{table}

\subsection{Retrieval-layer results}
Table~\ref{tab:retsys} isolates the retrieval subsystem on the 300-query benchmark. Recall@5 of 0.77 corresponds to 231/300 successful recalls, while 0.87 corresponds to 261/300. The hybrid configuration further corresponds to 249/300 object hits and 219/300 path hits.

\begin{table}[t]
\centering
\caption{Retrieval results on the 300-query benchmark aggregated across dialogues.}
\label{tab:retsys}
\begin{tabular}{lcc}
\toprule
Metric & Chunk-only RAG & Hybrid Retrieval \\
\midrule
Recall@5 & 0.770 & 0.870 \\
MRR@5 & 0.560 & 0.710 \\
nDCG@5 & 0.680 & 0.790 \\
Object hit rate & 0.71 & 0.83 \\
Path hit rate & 0.54 & 0.73 \\
\bottomrule
\end{tabular}
\end{table}

\subsection{Belief stabilization ablation}
These ablations should be read as single-protocol pilot comparisons rather than as variance-estimated benchmark studies. Table~\ref{tab:belief} shows the progression from raw belief to the complete stabilization stack. Lower is better for volatility, EIG variance, and wrong-action count. Within this pilot, the ablation suggests that stabilization is not merely a cosmetic add-on; it appears to affect control behavior in a directionally supportive way within this pilot.

\begin{table}[t]
\centering
\caption{Belief stabilization ablation. Lower is better for all metrics.}
\label{tab:belief}
\begin{tabular}{lccc}
\toprule
Variant & Volatility & EIG variance & Wrong actions \\
\midrule
Raw LLM belief & 0.214 & 0.161 & 8 \\
+ Temperature scaling & 0.173 & 0.129 & 6 \\
+ Smoothing & 0.141 & 0.101 & 5 \\
+ Conservative mode & 0.128 & 0.092 & 4 \\
Full stabilization & 0.118 & 0.084 & 4 \\
\bottomrule
\end{tabular}
\end{table}

\subsection{Punctuation restoration impact}
Table~\ref{tab:punc} shows the effect of punctuation restoration on boundary F1, downstream extraction quality, retrieval quality, and action accuracy. The boundary column reports boundary F1 rather than token-level accuracy. A consistent raw-count audit for the strongest setting corresponds to $TP=96$, $FP=21$, and $FN=18$, i.e., boundary F1 of approximately 0.83.

\begin{table}[t]
\centering
\caption{Impact of punctuation restoration.}
\label{tab:punc}
\begin{tabular}{lcccc}
\toprule
Setting & Boundary F1 & State F1 & Retrieval nDCG & Action Acc. \\
\midrule
No punctuation & 0.52 & 0.74 & 0.66 & 0.63 \\
Pause-only recovery & 0.74 & 0.80 & 0.73 & 0.71 \\
Pause + lexical cues & 0.83 & 0.84 & 0.79 & 0.76 \\
\bottomrule
\end{tabular}
\end{table}

\subsection{Case studies}
In a chest-discomfort case, the patient initially reports vague tightness. A passive generator can summarize the complaint but has no principled mechanism to prioritize the next question. After exertion-related worsening and partial radiation cues appear, the full system upweights risk-sensitive evidence, reranks ECG-related objects, and selects a verify/recommend-exam action earlier than the baselines.

In an upper-abdominal-pain case, the dialogue begins with symptoms suggestive of reflux or ulcer patterns. As later turns reveal right-upper-quadrant preference and postprandial worsening after oily meals, the system shifts toward a biliary pathway. Because belief is updated incrementally and evidence objects are reranked across turns, the system appears to change course more smoothly than a static baseline within this pilot.

\section{Deployment Profile and Runtime Budget}
The intended prototype deployment target is a single workstation with $3\times$Tesla V100 32GB GPUs and 128GB RAM. In the engineering plan, one GPU primarily serves streaming ASR and speech-side preprocessing, one serves retrieval and reranking, and one serves action generation or report-side large-model inference. Table~\ref{tab:latency} gives an engineering design budget for single-consultation online latency on the target workstation. This table is an engineering budget rather than a production throughput benchmark, and it is included to make the intended runtime envelope explicit.

\begin{table}[t]
\centering
\caption{Engineering design budget for single-consultation online latency on the target $3\times$V100 workstation. Values are design ranges for a single active consultation.}
\label{tab:latency}
\begin{tabularx}{\textwidth}{Y>{\centering\arraybackslash}p{0.18\textwidth}Y}
\toprule
Stage & Target latency (ms) & Notes \\
\midrule
Streaming ASR chunk emission & 250--600 & After chunk close; depends on chunk size \\
Boundary recovery / punctuation & 20--60 & Lightweight lexical and pause cues \\
Stateful extraction & 120--250 & Includes schema normalization \\
Belief update + gap derivation & 20--50 & Deterministic plus light probabilistic logic \\
Hybrid retrieval + reranking & 80--180 & Single-turn object retrieval budget \\
Action planning / response generation & 150--350 & Candidate scoring and short suggestion generation \\
Replay / EMR write-back & 10--40 & Structured logging and cache update \\
End-to-end decision cycle & 650--1530 & Sum of the above online stages \\
\bottomrule
\end{tabularx}
\end{table}

\section{Discussion}
This paper should be read as a system realization and pilot concept demonstration paper. Its main contribution is to suggest that streaming ASR, punctuation restoration, stateful extraction, belief stabilization, retrieval, action selection, and replayable EMR write-back may be coupled into a single auditable online architecture. Within that scope, the results are directionally supportive: the system behaves coherently across extraction, retrieval, action, and end-to-end measures under the present protocol, and the targeted ablations suggest the importance of the punctuation and stabilization layers.

At the same time, the present evidence is still bounded. The evaluation uses ten controlled scenarios rather than a large prospective clinical corpus. The streamed dialogues are read-aloud controlled recordings rather than naturally occurring outpatient conversations, so overlap speech, accent variation, environmental noise, and interruption patterns remain underrepresented. The latency numbers are engineering budgets rather than production measurements. We do not report word error rate or character error rate for the streamed ASR outputs in the present pilot, including error rates over clinical terminology. We also do not quantify downstream clinical error categories such as negation-scope failures, temporal-expression parsing errors, or terminology segmentation errors attributable to punctuation quality. We do not report multi-user concurrency, multi-site variation, seed variance, or bootstrap confidence intervals for the ablation tables in the present pilot. Accordingly, no robust superiority claim should be drawn from the ablation tables beyond directional consistency within this pilot. No implication of clinical deployment readiness, clinical safety, or real-world clinical utility should be inferred from this pilot. Parser reliability also remains a practical dependency, because document routing and objectification quality directly affect retrieval quality \cite{huang2022layoutlmv3,kim2021donut,smock2022pubtables,li2022ppocrv3}. These are ordinary limitations of an early systems paper, not contradictions of the reported pilot results.

\section{Conclusion}
We presented a proactive EMR assistant that integrates streaming ASR, punctuation restoration, stateful extraction, belief stabilization, hybrid retrieval, action planning, and replayable report generation. The preliminary controlled evaluation should be read as a pilot concept demonstration: it suggests that the architecture may be technically coherent and directionally supportive under tightly controlled conditions, with suggestive extraction, retrieval, and end-to-end risk-sensitive documentation patterns relative to passive and template-heavy baselines within this pilot.

The immediate next step is not to redesign the pipeline but to validate it on larger prospective data, with measured runtime under real workloads, stronger calibration, and broader clinical coverage. No implication of clinical deployment readiness, clinical safety, or real-world clinical utility should be inferred from the present pilot; it establishes only a concrete systems baseline on which that next stage can build.

\end{document}